\documentclass[runningheads]{llncs}

\usepackage{amssymb}
\usepackage{graphicx}
\usepackage{url}
\usepackage{color}
\usepackage{amsfonts}
\usepackage{blindtext}
\usepackage{mathrsfs}
\usepackage{bbm}
\usepackage{url}
\usepackage{algorithm}
\usepackage{algorithmic}
\usepackage{paralist}
\usepackage[english]{babel}
\usepackage{booktabs}
\usepackage{multicol}
\usepackage{multirow}
\usepackage{subfig}

\begin{document}
\title{Automatic Dialogic Instruction Detection for K-12 Online One-on-one Classes}
%
%
\author{Shiting Xu \and
Wenbiao Ding \and
Zitao Liu\thanks{Corresponding Author: Zitao Liu}}

\authorrunning{S. Xu et al.}
%
\institute{TAL Education Group, Beijing, China \\
\email{\{xushiting, dingwenbiao, liuzitao\}@100tal.com}}
\maketitle              
\begin{abstract}
Online one-on-one class is created for highly interactive and immersive learning experience. It demands a large number of qualified online instructors. In this work, we develop six dialogic instructions and help teachers achieve the benefits of one-on-one learning paradigm. Moreover, we utilize neural language models, i.e., long short-term memory (LSTM), to detect above six instructions automatically. Experiments demonstrate that the LSTM approach achieves AUC scores from 0.840 to 0.979 among all six types of instructions on our real-world educational dataset.
\end{abstract}

\section{Introduction}
\label{sec:intro}
With the recent development of technology such as digital video processing and live streaming, various forms of online classes emerge \cite{ChinaEducationResources2012}. Because of the better accessibility and live learning experience, one-on-one class stands out where students are able to not only study materials at their only own pace, but have opportunities to frequently interact with their teachers facially and vocally \cite{chen2019multimodal,li2020multimodal,10.1145/3366423.3380018}. Online one-on-one class has demonstrated its personalized education experience as supplements to the traditional training from public schools \cite{liang2005few}. 

In spite of the above benefits, online one-on-one classes pose numerous challenges on instructors. On one hand, the instructor qualifications are significantly different from those in public schools. Public school teachers focus on making sure that the majority students are on track and pass their qualification examinations. While one-on-one instructors need to pay detailed attentions to every single student and adjust their teaching paces, styles, or even contents accordingly. Furthermore, students enroll in one-on-one courses for high-frequency interactions. This requires the teachers to encourage and lead students' active participations. On the other hand, a large portion of one-on-one participants are academically low-ranking K-12 students. Most of them are eager to study but don't know how to learn. The one-on-one instructors are responsible to help them build effective study habits. Therefore, in order to scale the qualified supply of one-on-one instructors and provide more effective and personalized education to the general K-12 students, we develop six in-class dialogic instructions for one-on-one class teachers. Moreover, we build an end-to-end system to automatically detect and analyze the proposed pedagogical instructions.

\vspace{-0.1cm}
\section{Related Work}
\vspace{-0.1cm}
\label{sec:related}
Many existing methods have been developed to analyze classroom dialogic instructions. Wang et. al  identify teacher lecturing, class discussion and student group work in the traditional classroom by asking teachers to wear the LENA system \cite{ganek2016language} during the class \cite{wang2014automatic}. Donnelly et al. identify occurrences of some key instructional segments, such as Question \& Answer, Supervised Seatwork, etc., by using Naive Bayes models \cite{donnelly2016automatic}. Owens et al. develop Decibel Analysis for Research in Teaching, i.e., DART, to analyzes the volume and variance of classroom recordings to predict the quantity of time spend on single voice (e.g., lecture), multiple voice (e.g., pair discussion), and no voice (e.g., clicker question thinking) activities \cite{owens2017classroom}. 

Our work is distinguished from existing research studies because (1) we focus on the K-12 online one-on-one domain and propose six pedagogical instructions explicitly designed for it; (2) our dialogic instruction detection approach is an end-to-end solution that doesn't require any human intervention or any additional recording device. 

\vspace{-0.1cm}
\section{Our Approach}
\vspace{-0.1cm}
\label{sec:method}
\subsection{Dialogic Instructions}
\label{sec:dialogic}

By analyzing thousands of online one-on-one class videos and surveying hundreds of instructors, students, parents and educators, we categorize six dialogic instructions for K-12 online one-on-one classes as follows:

\begin{itemize}
\item \emph{greeting}: Greeting instructions help teachers manage their teaching procedures before the class, such as greeting students, testing teaching equipments. Examples: ``How are you doing?'', ``Can you hear me?'', etc.

\item \emph{guidance}: Guidance instructions ask teachers to interact with students when lecturing on a particular knowledge point or a factual answer. Examples: ``Do you know the reason?'', ``Let's see how we can get there?', etc.

\item \emph{note-taking}: Note-taking instructions require teachers to help students learn how to take notes and assist them to build effective learning habits. Examples: ``Highlight this paragraph.'', ``Please copy this part'', etc.

\item \emph{commending}: Commending instructions ask teachers to encourage students and build their confidence. Examples: ``Good job.'', ``Well done.'', etc.

\item \emph{repeating}: Repeating instructions remind teachers to let students retell the content by themselves, which enhances their understandings. Examples: ``Could you please explain that to me?'', ``Can you rephrase that?'', etc.

\item \emph{summarization}: Summarization instructions ask teachers to summarize teaching contents and materials at the end of the each class and conclude the main takeaways. Examples: ``Let's review the key points'', ``Let's wrap up.'', etc.

\end{itemize}

\subsection{The Dialogic Instruction Detection Approach}

\begin{figure}
\centering
\includegraphics[width=\textwidth]{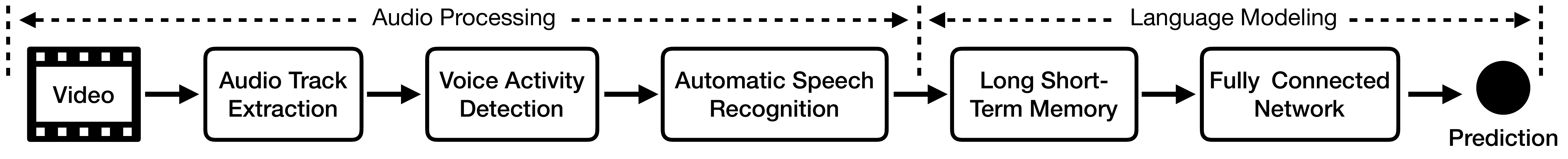}
\caption{The workflow of the end-to-end dialogic instruction detection approach.}
\label{fig:workflow}
\end{figure}

The end-to-end dialogic detection pipeline takes class recordings as input and outputs spoken sentences of the above six types of dialogic instructions. The entire workflow is illustrated in Figure \ref{fig:workflow}, which consists of two key components: \emph{Audio Processing} and \emph{Language Modeling}.

\vspace{-0.5cm}
\vspace{-1.2cm}

\subsubsection{Audio Processing} Audio processing involves three key steps: (1) extracting audio tracks from video recordings; (2) cutting audio tracks into short-span segments and removing noises and silence segments by a voice activity detection (VAD) algorithm; and (3) transcribing each audio segment by using an automatic speech recognition (ASR) algorithm. Please note that since both students' and teachers' videos are recorded separately, voice overlaps don't exist in the video recordings. This avoids the unsolved challenge of speaker diarization \cite{tranter2006overview,anguera2012speaker}.

\vspace{-0.5cm}
\subsubsection{Language Modeling} We conduct language modeling on the transcriptions from the audio processing module. For each word, we first fetch its low dimensional embeddings from a pre-trained word2vec model. After that, we build neural classifiers for each type of dialogic instructions defined in Section \ref{sec:dialogic}. In this work, we use the long short-term memory (LSTM) as our language modeling networks \cite{hochreiter1997long,greff2016lstm}. The LSTM models take a sentence as input and sequentially update the hidden state representation of each word by using a well designed memory cell, which is able to capture the long range dependencies within each sentence. The details of LSTM can be found in \cite{hochreiter1997long,greff2016lstm}. LSTM model have been successful in language modeling tasks such as text classification \cite{lai2015recurrent,yang2016hierarchical}, machine translation \cite{wu2016google}, etc. Finally, we build a two-layer fully-connected position-wise feed forward network on the last hidden representation of LSTM to conduct the final predictions.

\vspace{-0.1cm}
\section{Experiments}
\label{sec:experiment}
In this work, we collect 2940 sentences for each type of dialogic instruction by manually annotating class recordings from a third-party online one-on-one learning platform\footnote{https://www.xes1v1.com/}. Each sentence is associated with a binary label, indicating whether the sentence belongs to a dialogic instruction. We use 2352 sentences for training and the rest for validation and testing. Similar to Blanchard et al. \cite{blanchard2015study}, we find that publicly available AI engines may yield inferior performance in the noisy and dynamic classroom environments. Therefore, we train our own VAD \cite{tashev2016dnn}, ASR \cite{zhang2018deep} and word2vec \cite{mikolov2013distributed} models on the classroom specific datasets. 

We compare the LSTM language modeling network with several widely used baselines: logistic regression \cite{hosmer2013applied}, i.e, \emph{LR}, support vector machine \cite{suykens1999least}, i.e., \emph{SVM}, and gradient boosting decision trees \cite{friedman2002stochastic}, i.e., \emph{GBDT}. Similar to Tang et al. \cite{tang2018time}, we use area under curve (AUC) score to evaluate the model performance \cite{fawcett2006introduction}.

\vspace{-0.1cm}
\subsection{Model Performance}

\begin{figure}
\centering
\includegraphics[width=\textwidth]{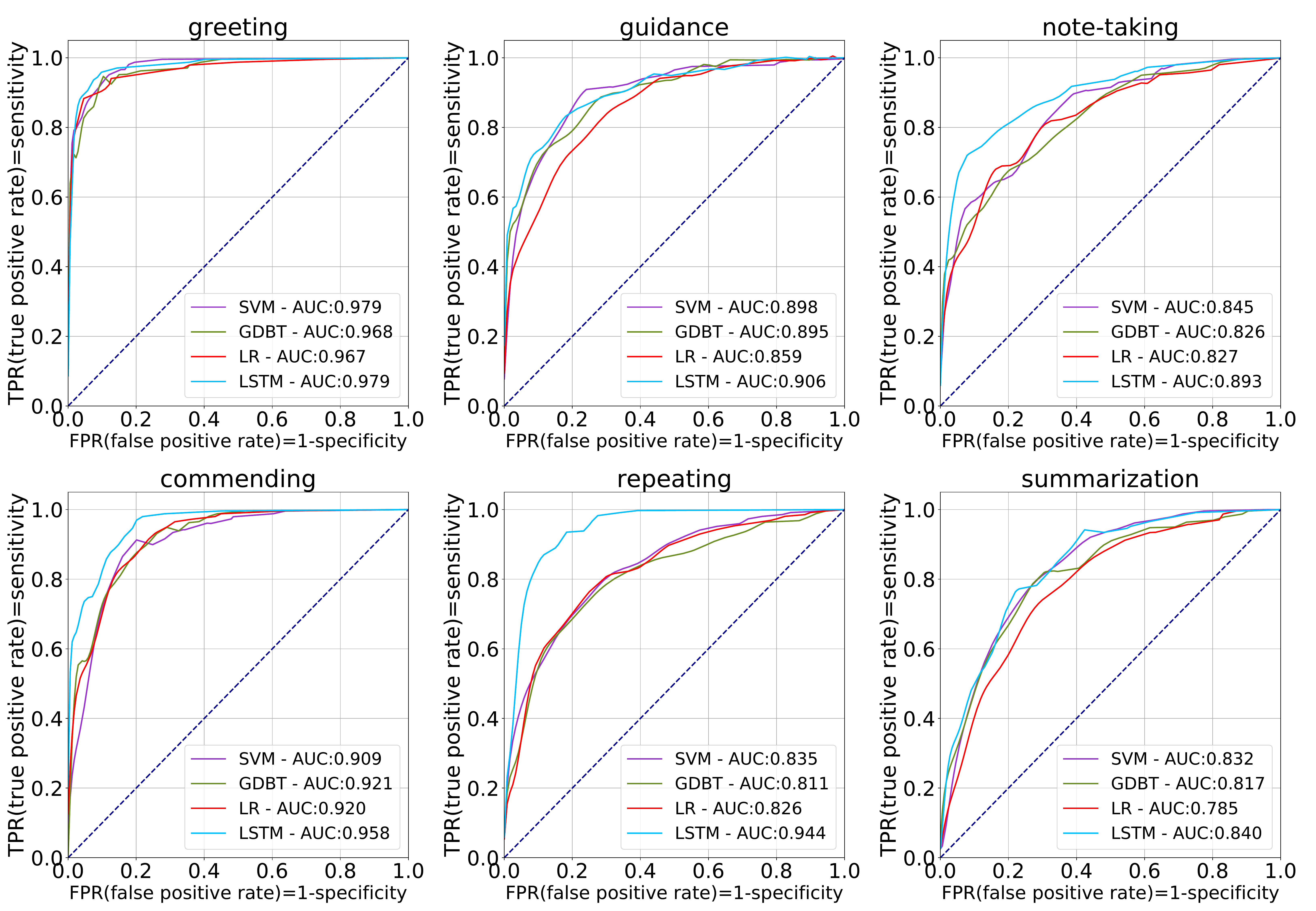}
\vspace{-0.5cm}
\caption{ROC curves for detection performance of six dialogic instructions.}
\label{fig:results}
\end{figure}

Figure \ref{fig:results} shows that our LSTM approach outperforms all other methods  on all six types of dialogic instruction detection tasks. Specifically, from Figure \ref{fig:results}, we find that simple instructions are relatively fixed and have little variants, such as ``greeting'' and ``summarization''. All the approaches have comparable performance. While for complex instructions with many language variations such as ``note-taking'', ``commending'' and ``repeating'', \emph{LSTM} significantly outperforms other baselines by large margins. We believe this is because the sequential neural networks are able to capture the long contextual language dependence within the sentence, which is very important when dealing with colloquial conversations.

\vspace{-0.7cm}
\vspace{-1.1cm}

\section{Conclusion}
\label{sec:conclusion}
In this work, we propose six dialogic instructions and build an end-to-end solution for online one-on-one instructors. Experiments on a real educational dataset show that our LSTM based approach outperforms other baselines in the proposed six dialogic instructions.

%
%
%
%
\bibliographystyle{splncs04.bst}
\bibliography{aied2020}
\end{document}